# Decision Transformer for Enhancing Neural Local Search on the Job Shop Scheduling Problem

Constantin Waubert de Puiseau, Fabian Wolz, Merlin Montag, Jannik Peters, Hasan Tercan, and Tobias Meisen

*Abstract*— **The job shop scheduling problem (JSSP) and its solution algorithms have been of enduring interest in both academia and industry for decades. In recent years, machine learning (ML) is playing an increasingly important role in advancing existing and building new heuristic solutions for the JSSP, aiming to find better solutions in shorter computation times. In this paper we build on top of a state-of-the-art deep reinforcement learning (DRL) agent, called Neural Local Search (NLS), which can efficiently and effectively control a large local neighborhood search on the JSSP. In particular, we develop a method for training the decision transformer (DT) algorithm on search trajectories taken by a trained NLS agent to further improve upon the learned decision-making sequences. Our experiments show that the DT successfully learns local search strategies that are different and, in many cases, more effective than those of the NLS agent itself. In terms of the tradeoff between solution quality and acceptable computational time needed for the search, the DT is particularly superior in application scenarios where longer computational times are acceptable. In this case, it makes up for the longer inference times required per search step, which are caused by the larger neural network architecture, through better quality decisions per step. Thereby, the DT achieves state-of-the-art results for solving the JSSP with ML-enhanced search.**

## I. INTRODUCTION

Effective and efficient scheduling presents an ongoing challenge that is critical for the success in many sectors, from manufacturing [1, 2] to cloud computation [3]. Scheduling, broadly, deals with the allocation of resources to tasks over time with the goal of optimizing a given objective [4]. The job shop scheduling problem (JSSP) is an abstracted combinatorial scheduling problem that underlies many real-world problems and has been extensively studied in the literature. To this day, new solution methods for scheduling problems are developed and tested on the JSSP due to its interesting properties and the availability of popular public datasets that allow for rigorous benchmarking [5–7].

In recent years, machine learning has emerged as a promising technique for new solution methods for scheduling problems by enhancing or replacing heuristic decisions in existing dispatching rules [8–12], metaheuristics and search heuristics [13–16], and optimal solvers [17] to varying degrees. A notable advancement in this area is the integration of deep reinforcement learning (DRL) with local search algorithms. For instance, Falkner et al. [18] proposed an approach wherein a DRL agent is trained to learn three critical components of a local search algorithm: solution acceptance, neighborhood operators, and perturbation decisions. This ML-enhanced search significantly outperformed state-of-the-art search methods. Meanwhile, transformer architectures [18] are nowadays deployed at scale and achieve breakthrough results in various domains, most publicly known in natural language processing [19], by capturing and processing important features in sequential data. Aiming at harnessing the strength of sequential data processing of transformers, decision transformers (DTs) have recently been introduced for learning strategies in sequential decision-making processes from simulation data [20].

In this work, we integrate the DT with the NLS approach by learning from data generated during the local search process of trained NLS models. Using the DT in combination with NLS comes with two potential benefits over the original NLS algorithm: First, by considering a history of past actions, the DT can learn how previous decisions have influenced the current problem instance and adjusts its strategy accordingly. Second, the DT is trained to align its actions with a manually set reward prior. This can lead to performance improvements over teacher algorithms – in our case, the already highly competitive NLS models.

The main contributions of this paper are:

- The introduction of a decision transformer approach to boost the performance of a learned local search heuristic on the JSSP.
- An experimental analysis of both the performance and the learned behavior of the model with respect to the teacher models.

The remainder of this work is structured as follows: first, the JSSP and general solution approaches are introduced in section II. Here, we also provide a brief introduction to DRL, DTs, and the neural neighborhood search algorithm that jointly build the basis of our work. Next, related work is discussed in section III before our methods are detailed in section IV. We present the achieved results in section V, and a detailed comparison between the teacher and DT in terms of practical implications and learned behavior is provided in section VI. Section VII gives a conclusion and an outlook for future work.

*Corresponding author: C. Waubert de Puiseau. All authors are with the* Institute for Technologies and Management of Digital Transformation, Wuppertal, Germany (emails: {waubert, fabian.wolz, merlin.montag, jpeters, tercan, meisen}@uni-wuppertal.de).

## II. PRELIMINARIES

### A. Job Shop Scheduling Problems

The classical, abstract version of the JSSP considered in this paper deals with the allocation of a set of i jobs $J_i$, each comprising j operations $O_j$ with processing times $P_{ij}$, to m machines $M_m$ over time, with the objective of minimizing the maximum completion time $C_{max}$ - that is, the duration from the initiation of the first operation to the completion of the last one, commonly referred to as the makespan. Operations within each job underlie precedence and no-overlap constraints, such that any job visits each machine exactly once in a predefined order. Accordingly, the number of operations j and the number of machines m is equal in this setup. In addition, each machine may only process one job at a time. In terms of notation, we refer to problem instances with j jobs and m machines as instances of size j x m. For example, a 20x15 problem instance consists of 20 jobs, each with 15 operations on 15 machines in total. The described JSSP problem is NP-hard, making enumerative algorithms for finding optimal solutions to large instance sizes impractical in real-world applications.

### B. Solution Methods for Job Shop Scheduling Problems

Solutions algorithms for the JSSP have been under active development and research for decades in the operations research domain. These algorithms are designed to achieve a balance among minimizing the objective (e.g., makespan), ensuring rapid execution times, and requiring minimal implementation effort.

Provenly optimal solutions can be found by means of constraint programming solvers such as the CP-SAT solver by Google OR-Tools [21]. However, such solvers become impractical for large JSSP instances due to the exponential growth of the solution space that must be explored.

A common category of solution methods in practice are priority dispatching rules (PDRs), in which dispatching and ordering decisions are governed by job prioritization rules such as “shortest processing time first”. A representative collection of such rules has been studied by Haupt et al. [22]. Although PDRs are straightforward to understand and implement in industrial settings, they often fail to produce competitive makespans when compared to other solution methods.

Another solution category comprises a variety of (meta-) heuristics, such as the shifting bottleneck heuristic [23], genetic algorithms [24] and heuristic search algorithms [25, 26]. Depending on the problem size and considered constraints, tailored metaheuristics have long been the de-facto state-of-the-art.

With the advent of more powerful machine learning algorithms, particularly deep learning techniques, the aforementioned solution methods have increasingly been either enhanced or partially replaced by data-driven approaches in recent years. Further details and significant developments related to this work are discussed in Section III, “Related Work.”

### C. Deep Reinforcement Learning

DRL is a machine learning paradigm, in which an agent learns a parameterized policy, approximated by a deep neural network, through interaction with an environment. In each timestep t, the agent receives a representation $S_t$ of the state of the environment, takes an action $A_t$ which alters the state of the environment and receives a feedback signal $R_t$, called reward. The training process aims at maximizing the cumulated reward across an episode, $\sum_{t_0}^{t_{final}} r_t$, i.e., from the first interaction at $t_0$ to the final interaction $t_{final}$. A common way to learn a policy is using the Q-Learning algorithm. Q-values, or state-action-values, are assigned to each state-action pair and represent the total cumulative reward that can be achieved by taking action $A_t$ in state $S_t$ and subsequently following the learned policy. Once Q-values have been learned from experience, the optimal policy is derived by iteratively estimating the Q-values for all possible actions in each state of the episode and selecting the action that corresponds to the highest predicted Q-value.

DRL can be applied to any sequential decision-making process that adheres to the Markov property, meaning that the state of the environment at any given time is independent of prior states. [27]

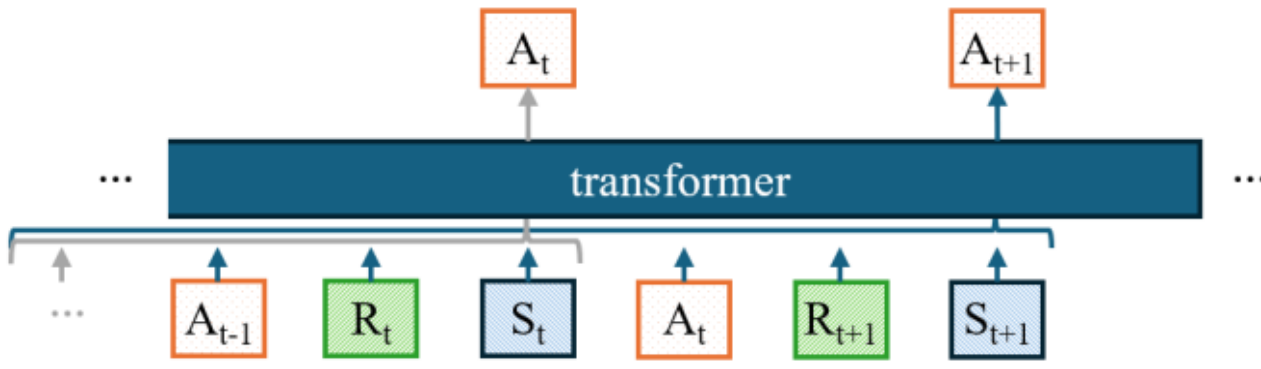

**Fig. 1**: Depiction of the Decision Transformer and its inputs, adapted from [20]

### D. Decision Transformers

The DT is an architecture for offline reinforcement learning, in which Markov decision processes are abstracted as sequence-modelling problems [20]. The DT has already been successfully applied in various domains such as active object detection in robotics [28], recommender systems [29] and chip placement optimization [30]. In traditional RL, models output (or evaluate) actions A given a representation of the current problem state *S* and receive a reward *r* in response. Hence, these sequences may be represented as lists of collected state-action-reward tuples, [{$S_0$, $A_0$, $r_0$}, {$S_1$, $A_1$, $r_1$} … {$S_{final}$, $A_{final}$, $r_{final}$}]. In contrast, DTs are trained to output actions by considering as input the sequence of the last $K$ states, actions, and the corresponding return-to-go values $R$. In this context, $K$ is referred to as the context length. The return-to-go at a given time step $t$, denoted as $R_t$, is defined as the sum of the remaining rewards: $R_t = r_t + r_{t+1} + \cdots + r_{T-1} + r_T$. At each time step *t*, the DT uses the sequence of past states, actions, and returns-to-go, along with the current state $S_t$ and return-to-go $R_t$. This is formally expressed as follows:

$$A_t = DT(R_{t-k}, S_{t-k}, A_{t-k}, \ldots, .R_t, S_t). \quad (1)$$

In other words, the DT learns to generate action sequences that aim to achieve the manually set target R (i.e., the prior) by minimizing the difference between R and the cumulated sum of rewards, instead of minimizing the cumulated sum or rewards directly. The underlying concept is depicted in **Fig. 1**.

The mapping function of the DT is approximated with a transformer architecture, originally designed for processing sequential data in text format [18]. In the case of DTs, instead of embedding words into tokens, we embed returns-to-go, states and actions into a fixed vector space. For more details on the DT, readers are referred to the work by Chen et al. [20]. We give details on our own implementation in section IV.

### *E. Neural Local Search*

NLS is a recent approach to control a local neighborhood search heuristic on the JSSP with a DRL agent [31]. The underlying local search (LS) heuristic first constructs an initial solution and then takes iterative steps to improve upon it. In each iteration, the heuristic:

- accepts or declines the solution of the last iteration,
- chooses a new neighborhood operation that defines the next neighborhood, i.e. the set of solutions to integrate in the search,
- or chooses a perturbation operator to jump to a new area in the search space in which to continue the search.

Falkner et al. [31] translate the heuristic decisions into actions that may be taken by a DRL agent, experimenting with three different action spaces that are defined by the following three sets of actions:

1. Acceptance decisions: the decision whether the last LS step is accepted or not:
$$A_A \; := \; \{0, 1\}$$
2. Acceptance-Neighborhood decision: tuple representing the above acceptance decision and the choice between four different neighborhood operations in the set $\Phi$:
$$A_{AN} \; := \; \{0, 1\} \, x \; \Phi$$
3. Acceptance-Neighborhood-Perturbation decision: tuple representing the acceptance and neighborhood decisions plus a perturbation decision from the perturbation operator set $\Psi$:
$$A_{ANP} \; := \; \{0, 1\} \, x \, \{ \Phi \cup \Psi \}$$

The neighborhood operators comprise: the Critical Time (CT) operator, which swaps adjacent nodes in critical blocks [32], the Critical End Time (CET) [33] operator, which swaps nodes at the start or end of a critical block, the Extended Critical End Time (ECET) [34] operator, which swaps nodes at both the start and end of a critical block, and the Critical End Improvement (CEI) [35] operator, which shifts a node within a critical block to a new position within the same block.

All actions are taken based on states that are derived from a learned representation consisting of the current problem instance to be solved and its current solution. Specifically, the authors employ an encoder-decoder architecture depicted in **Fehler! Verweisquelle konnte nicht gefunden werden.**. The encoder generates a representation of the problem instance and its current solution using a graph neural network, which is then enriched with the following explicit state features: the current makespan, the best makespan found so far, the last acceptance decision, the last applied operator, the current time step, the number of consecutive steps without improvement, and the number of perturbations applied so far. Utilizing this enriched representation, the decoder computes Q-values that guide the selection of the next action. These output Q-values are learned with Double Deep Q-Learning [36] aiming to maximize the total improvement in makespan between the initial and final solution across the episode.

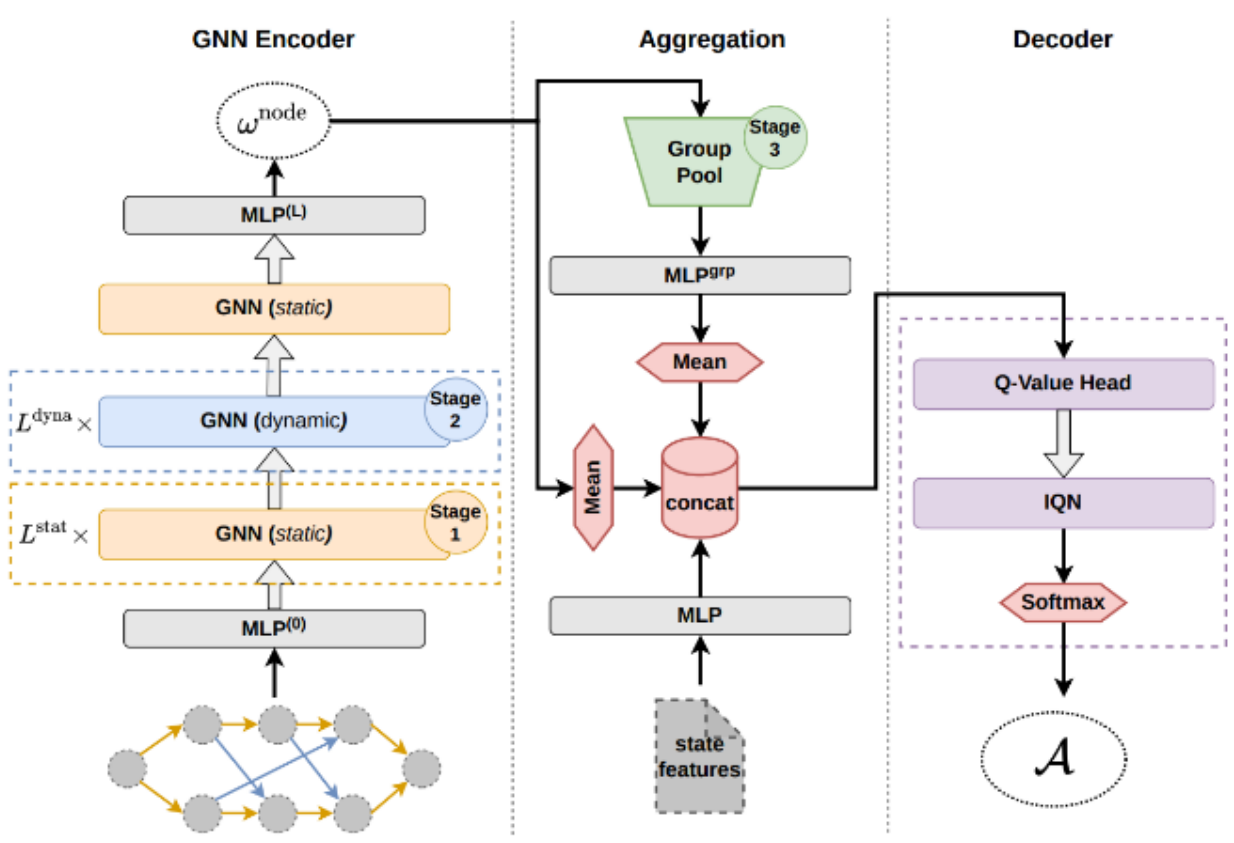

**Fig. 2**: Neural network architecture of the NLS approach, reprinted from [31]

To this end, Falkner et al. [31] proposed a dense clipped reward that returns the makespan difference before and after the local search step induced by the action:

$$r = max(m_{t-1} - m_t \, , 0)$$

## III. Related Work

### *A. Learned Construction Heuristics*

A plethora of deep learning based solution methods to the JSSP have been proposed in recent years [11, 37–42], many of which serve as learned construction heuristics. Construction heuristics iteratively generate a solution by starting with an empty schedule and, in each iteration, determining the next unscheduled operation of a job to be added to the schedule. Learned construction heuristics have shown very promising results for the JSSP, showcasing that deep learning models are capable of capturing the structure of the JSSP effectively. In this study, we refrain from applying the DT to construction heuristics for two reasons: firstly, although they perform well on the vanilla JSSP, the transfer to scheduling scenarios with additional constraints remains an open problem, where search heuristics are still dominant [14, 41]. Secondly, one of the advantages of the DT is that it can effectively take past search actions on the same instance into consideration. This strength cannot be leveraged for construction heuristics, in which all relevant information of past actions is summarized in the partially solved schedule, which cannot be altered retrospectively.

### *B. Learned Heuristic Search*

In contrast to construction heuristics, in learned heuristic search, certain decision rules of search algorithms are replaced by neural network inferences. Typically, the search algorithms aim at improving an initial solution. Though the reported results on the JSSP are slightly worse than those of learned construction heuristics, learned heuristic search algorithms are more easily adaptable to problems with more real world constraints and objectives, because their action spaces remain constant [14]. This property makes them worth investigating, considering that most real-world use-cases are subject to multiple constraints.

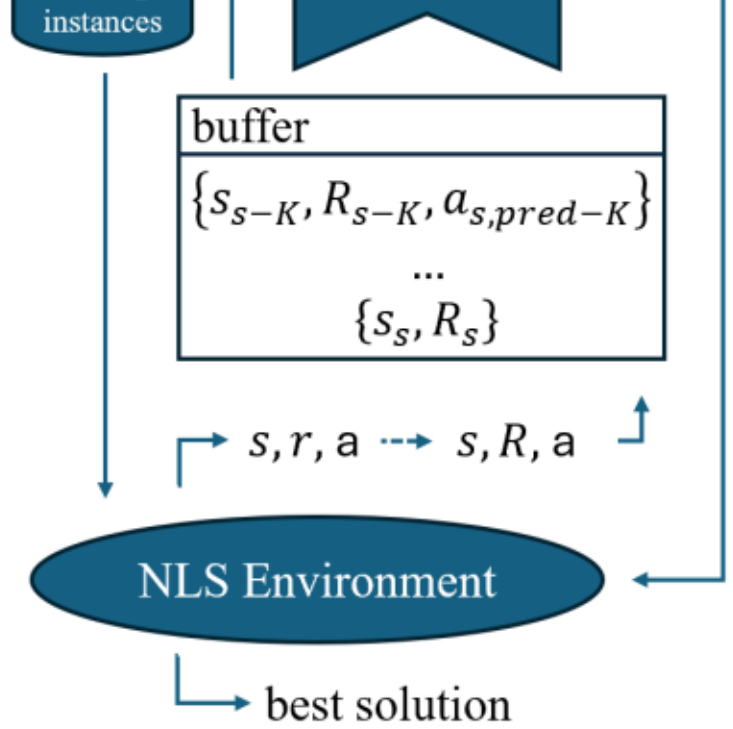

**Fig. 3**: Schematic depiction of the three steps for the DT development

A well-known representative of this category is local rewriting, in which agents iteratively decide on the local region and rewriting strategy for that region [16]. Another noteworthy approach was demonstrated by Reijnen et al., who train an agent to control parameters of an evolutionary algorithm on a multi-objective flexible JSSP [14].

In this study, we build upon the most recent state-of-the-art approach by Falkner et al. [31], in which an agent controls acceptance, neighborhood and perturbation decisions, as described earlier in section II.

### C. Imitation Learning on the JSSP

In imitation learning scenarios, “student”-models learn to imitate the behavior in sequential decision making of a “teacher” in a supervised manner by following labels generated by the teacher instead of rewards. In the JSSP, teacher algorithms can range from dispatching rules to exact algorithms. The goal of this method is to imitate the behavior of the teacher and then either surpass it (e.g., a dispatching rule) or achieve faster computational inference times (e.g., compared to an exact algorithm). For example, Ingimundardottir and Runarsson trained a linear machine learning model on data generated with dispatching rules [43]. Rinciog et al., for a related scheduling problem, pre-trained a neural network on the earliest-due-date dispatching rule before fine-tuning it with DRL [44]. Although these approaches showed some success of imitation learning, the results are far from competitive in terms of the absolute results compared to other algorithms due to their weak teacher algorithms.

In contrast, Lee and Kim train a graph neural network on optimal solutions, but also do not reach a competitive performance [45]. Hypothesizing that optimal solutions provide noisy data from which it is difficult to learn, Corsini et al. [38] and Pirnay et al. [39] have instead very recently suggested to train a student model on its own most successful solutions generated during sampling. All imitation learning approaches mentioned here are either construction heuristics or online dispatchers. To the best of our knowledge, our study, by training a DT, is the first to address imitation learning for a learned heuristic search method.

## IV. Methods

In the following, we describe how the DT and NLS are conceptually combined into one trainable algorithm for the JSSP and which specific design choices were taken. The descriptions of our method are structured along the three necessary steps “Dataset Generation”, “DT-Training” and “DT-Testing”, which are conceptually visualized in **Fig. 3**.

### A. Dataset Generation

**Training Instances** To train the DTs in a supervised manner, we must create suitable labeled datasets. To this end, we randomly generate 1000 problem instances for each problem size in the Taillard benchmark dataset according to the same reported specifications [5]. We then collect the state-action-reward tuples from local search trajectories on these instances with 100 and 200 search iterations generated by the interaction of the teacher NLS models with the NLS environment. 100 and 200 are in the range of iteration steps investigated in the original publication [31]. The collected information form preliminary raw labeled datasets.

**Teacher Models** The teacher models are pretrained NLS checkpoint models, available online in the original repository [46], for each problem size. We do this such that the DTs can learn from the successful learned solution strategies of the teachers. Note that the input to the DT, unlike that for the teacher models, is not only the state but also the return-to-go. We therefore expect the DT to be able to surpass the teacher models’ performance during inference by giving a suitable (challenging) return-to-go prior. A small value would indicate that the DT should act to achieve a small makespan, while a large value should have the opposite effect.

Within the original NLS models which are suitable as teachers, specific action sets were most successful for different problem sizes.

TABLE 1
ACTION SET USED IN THIS PAPER PER RESPECTIVE PROBLEM SIZE. USED ACTION SETS MARKED WITH "X".

| | 15x15 | 20x15 | 20x20 | 30x15 | 30x20 | 50x15 | 50x20 | 100x20 |
|---|---|---|---|---|---|---|---|---|
| $A_A$ | - | - | - | - | - | - | - | - |
| $A_{AN}$ | - | X | X | X | - | - | - | - |
| $A_{ANP}$ | X | - | - | | X | X | X | X |

For example, action set $A_A$ showed the best performance for 15x15 and $A_{ANP}$ for 30x20 problems (cf. the original results of Falkner et al. [31] reprinted in TABLE 3). Aiming for best teacher model performances for the dataset creation, we use the best respective reported action sets for each problem size to generate training data. There are two exceptions to this rule, problem sizes 15x15 and 20x20. In these two cases, $A_A$ performs best, but preliminary experiments showed that the NLS models had learned to take the same action ("accept") exclusively, which resulted in DTs that simply overfit on the training data and returned the exact same results as their teacher models. The same phenomenon of overfitting to taking one action was observed in the 15x15 $A_{AN}$ teacher model. For this reason, we resort to the action sets in TABLE 1 for the teacher models per problem size throughout this paper:

**NLS Environments** The NLS environments are closely adapted from [31] and implement the corresponding action dynamics and evaluations. As previously mentioned, we perform local search for 100 and for 200 interactions with these environments and record the state-action-reward tuples to create raw labeled datasets. As in the original NLS publication [31], the initial solutions needed for NLS are obtained from the FDD/MWKR composite dispatching rule introduced by Sels et al. [47], combining information on the flow due date and remaining work of each job. The local searches result in two raw datasets per problem size, with 100,000 and 200,000 data points, respectively:

$$dataset_raw = [\ \underbrace{\{s_0, a_0, r_0\}^0}_{one\ raw\ datapoint},\quad \dots, \{s_i, a_i, r_i\}^t]$$

where $i \in [100, 200]$ for the number of iterations and $t \in [0, \dots, 1000]$ for all training instances.

The returns-to-go required for DT training are obtained by traversing the training data backwards and recursively adding the rewards within a local search sequence, such that the returns-to-go $R(s)$ per step become $R_s = \sum_{i-s}^{i} r_s$, and the final datasets have the following format:

$$dataset_final = [\ \underbrace{\{s_0, a_0, R_0\}^0}_{one\ final\ datapoint},\quad \dots, \{s_i, a_i, R_i\}^t]$$

*B. DT-Training*

**Training Procedure** The DT model is trained for 500 epochs to minimize the categorical cross entropy loss between the predicted action $a_{pred}$ and the label action $a$. Preliminary experiments with varying hyperparameters on the 15x15 JSSP indicated that a context length K between 50 and 200 was most effective. To balance effectiveness and model size $K = 50$ is chosen. The Adam optimizer [48] is used in combination with cosine learning-rate-decay to smoothly reduce the overall learning rate, with periodical increases, back to the initial value of $6e^{-2}$. Unlike the NLS model, the DT generates an action based on the last K state-action-reward tuples. Given that the dataset contains only single tuples, a buffer is used that returns the last K tuples depending on the current iteration step. For the iteration steps that are smaller than the context length K, the remaining tokens are zero-padded and masked within the self-attention mechanism. We test our model every 33 epochs on 30 JSSP instances of the respective size, generated with a different random seed than the training and evaluation instances, and save the weights whenever the achieved test makespan improved.

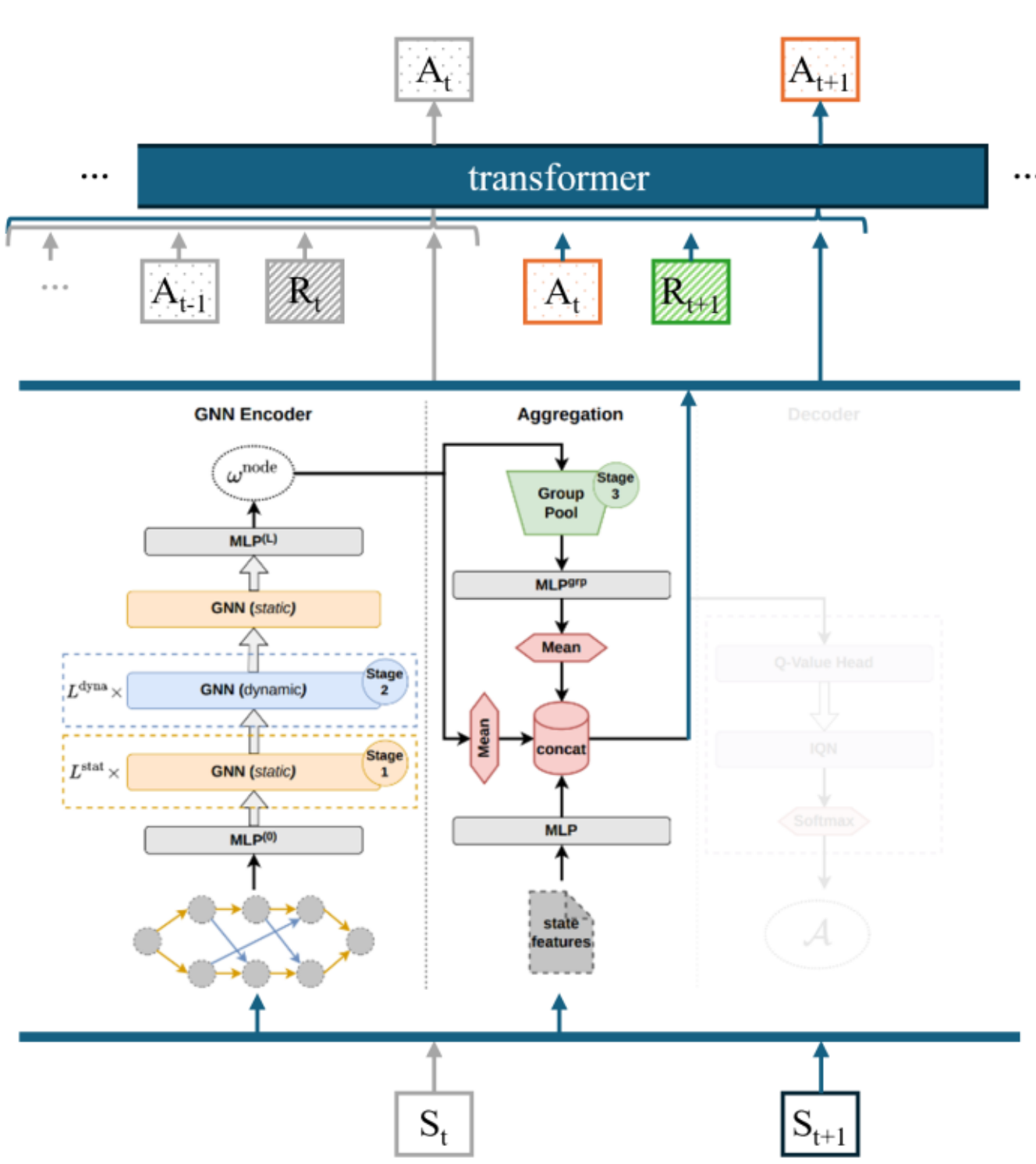

**Fig. 4**: Schematic view of the integration of the learned state space representation

**Model Architecture and Hyperparameters** Our architecture combines components of the original NLS model and our DT implementation. We use the learned latent space representation generated by the aggregator component in the NLS architecture as state embeddings. The original pre-trained checkpoint model weights, which were also used during data generation, were taken from [46].

Our DT transformer implementation is based on the minGPT model by Karpathy [49], from which most hyperparameters were adopted (see Table 2). The DT transformer replaces the decoder part of the original models to generate the action distribution in the output layer, as shown in **Fig. 4**. State embeddings, actions and rewards are all projected to the same embedding dimension of the tokens through linear layers. The resulting sequence of tokens is then fed into the first multi-head attention layer of the transformer. In contrast to the original implementation for textual data, our positional embedding is not created in relation to the block size but considers the maximum trajectory length of the problem, i.e., the number of iteration steps.

TABLE 2
HYPERPARAMETERS OF THE MINGPT MODEL OF THE DT

| Hyperparameter / Design | Value |
|---|---|
| Number of layers | 6 |
| Number of attention heads | 8 |
| Embedding dimension | 128 |
| Batch size | 512 |
| Context length K | 50 |
| Nonlinearity | GeLU |
| Dropout | 0.1 |
| Adam betas | (0.9, 0.95) |
| Grad norm clip | 1.0 |
| Weight decay | 0.1 |
| Learning rate decay | Cosine decay |

### *C. DT-Testing*

**Augmented Return-to-Go** As described earlier, the original NLS environment returns the clipped relative makespan improvement before and after the action is executed as reward signal, starting at step zero with an initially created schedule.

This presents a challenge during inference with the DT that requires a return-to-go value as input. Ideally, in step zero, the return-to-go $R_0$ would be the difference between the makespan of the initial schedule $m_{init}$ (obtained from the FDD/MWKR dispatching rule) and the makespan of the optimal solution $m_{optimal}$: $R_0 = m_{init} - m_{optimal}$. However, $m_{optimal}$ is usually unknown prior to solving it. Therefore, instead, we calculate a lower bound makespan $m_{lb}$, defined as the maximum sum of processing times p of operations $O_m$ that require the same machine:

$$m_{lb} = \max_{machine\ \epsilon\ M} \sum_{o \epsilon O_m} p(o)$$

Note that $m_{lb} \leq m_{optimal}$ in all cases since the $m_{lb}$ solution practically ignores precedence constraints within jobs. We then define our initial return-to-go as $R_0 = m_{init} - m_{lb}$ and reduce it in each step by the given reward. Since this $R_0$ is an optimistic approximation, it forces the DT to take the best possible actions.

## V. NUMERICAL RESULTS

### *A. Results on Taillard Benchmark*

TABLE 3 shows the results of NLS models with different action sets and the DTs on the Taillard benchmark dataset [5], which consists of ten instances for each of the reported instance sizes. In the table, $DT_{100}$ represents those DTs that were trained on the datasets in which only 100 search steps were performed by NLS. Nevertheless, all presented results are generated using 200 local search steps of the DT. The best results per column are printed in bold. Note that we re-printed the results by Falkner et al. [31] on those action sets that were not used as teacher models in gray.

It is evident that the DTs can outperform the NLS teacher models in almost all problem sizes and in some cases by significant margins. On average, using $DT_{100}$ leads to 1.11%p, 1.23%p and 1.15%p better optimality gaps than $NLS_{ANP}$, $NLS_{AN}$, $NLS_{A}$, respectively. Note that for the JSSP, such improvements are non-trivial. While DT and $DT_{100}$ perform similarly well on average, each taken alone does not consistently outperform the NLS models. In practice, this implies that both variants should be tried to achieve the best final result for a problem size.

### *B. Results on Own Test Instances*

To validate our observations on a larger set of instances, the results on 100 randomly generated JSSP instances are reported in TABLE 4. Indeed, the results on this larger test dataset confirm a similar trend that the DTs outperform NLS on average. However, compared to the results on the benchmark instances, they do so by a smaller margin and not as consistently. This indicates that the performance of each DT and $DT_{100}$ models varies between different problem instances.

## VI. STUDENT-TEACHER COMPARISON

In this section, we examine the learning behavior of the DT models in detail. Recall that they differ from their respective NLS teachers in two noteworthy ways.

Firstly, DT models can base decision on a context of up to 50 previously taken actions, states and returns-to-go. This can enable them to leverage knowledge about the influence of past decisions of their local search on the same problem instance. In fact, in preliminary experiments, we observed performance improvements only when contexts lengths of 50 steps or larger were used. On the downside, to take the context into account, a much larger neural network architecture is used which leads to increased inference times compared to the original NLS architectures. The second major difference to DRL models aiming at minimizing the cumulative rewards is that DT models have an artificial return-to-go prior which the models are trained to match, and which can be used to push them towards shorter makespans than those observed during training.

These differences lead to the following three questions we aim to answer in the following analysis to deepen our understanding of the proposed method:

1. What are the practical implications of using the comparatively larger and slower DT models during inference with respect to performance?
2. Is there a correlation between relatively better performance of DT models in comparison with their teacher models and a greater deviation in learned behavior?
3. Is there an optimal return-to-go to achieve the best performance?

The results in Table 1 and Table 2 indicate a better performance of the model when the same number of local search steps is performed. However, inferences of the DT take longer to compute than those of the teacher models on the same hardware.

TABLE 3: RESULTS (MAKESPANS AND OPTIMALITY GAPS) ON TAILLARD BENCHMARK INSTANCES [5]

| | | 15x15 | 20x15 | 20x20 | 30x15 | 30x20 | 50x15 | 50x20 | 100x20 | average |
|---|---|---|---|---|---|---|---|---|---|---|
| $NLS_A$ | gap | 7.74% | 12.16% | 11.54% | 14.33% | 19.42% | 18.00% | 11.22% | 5.89% | 12.54% |
| | makespan | 1324.0 | 1530.7 | 1804.0 | 2043.1 | 2267.0 | 3273.0 | 3163.0 | 5682.0 | 2635.85 |
| $NLS_{AN}$ | opt. gap | 8.72% | 11.39% | 11.67% | 14.31% | 19.57% | 18.29% | 11.15% | 5.84% | 12.62% |
| | makespan | 1336.0 | 1520.2 | 1806.0 | 2042.7 | 2270.0 | 3281.0 | 3161.0 | 5679.0 | 2636.99 |
| $NLS_{ANP}$ | opt. gap | 10.42% | 15.63% | 13.83% | 13.82% | **19.10%** | 10.62% | 10.83% | 5.73% | 12.50% |
| | makespan | 1357.0 | 1578.1 | 1841.0 | 2033.9 | **2261.0** | 3068.5 | 3152.0 | 5673.0 | 2620.56 |
| DT | opt. gap | 8.63% | **11.15%** | 11.73% | 14.12% | 19.21% | **10.55%** | **10.59%** | 5.88% | 11.48% |
| | makespan | 1335.0 | **1517.0** | 1807.0 | 2039.3 | 2263 | **3066.4** | **3145.0** | 5681.0 | 2606.71 |
| $DT_{100}$ | opt. gap | **7.66%** | 12.05% | **11.42%** | **13.62%** | 19.26% | 10.74% | 10.83% | **5.56%** | **11.39%** |
| | makespan | **1323.0** | 1529.3 | **1802.0** | **2030.4** | 2264.0 | 3071.7 | 3152.0 | **5664.0** | **2604.55** |
| optimal | makespan | 1228.9 | 1364.8 | 1617.3 | 1787.0 | 1898.4 | 2773.8 | 2843.9 | 5365.7 | 2359.98 |

TABLE 4: RESULTS (MAKESPANS) ON 100 RANDOMLY GENERATED INSTANCES

| | 15x15 | 20x15 | 20x20 | 30x15 | 30x20 | 50x15 | 50x20 | 100x20 | average |
|---|---|---|---|---|---|---|---|---|---|
| $NLS_{AN}$/$NLS_{ANP}$ | 1319.7 | 1499.4 | **1736.8** | 1972.2 | **2178.2** | 2974.9 | 3137.0 | 5694.7 | 2564.1 |
| DT | 1310.0 | **1496.4** | 1739.5 | **1972.1** | 2181.0 | **2974.7** | 3138.1 | 5689.6 | 2562.7 |
| $DT_{100}$ | **1301.3** | 1499.2 | 1737.0 | 1972.6 | 2181.1 | 2975.3 | **3136.9** | **5675.2** | **2559.8** |

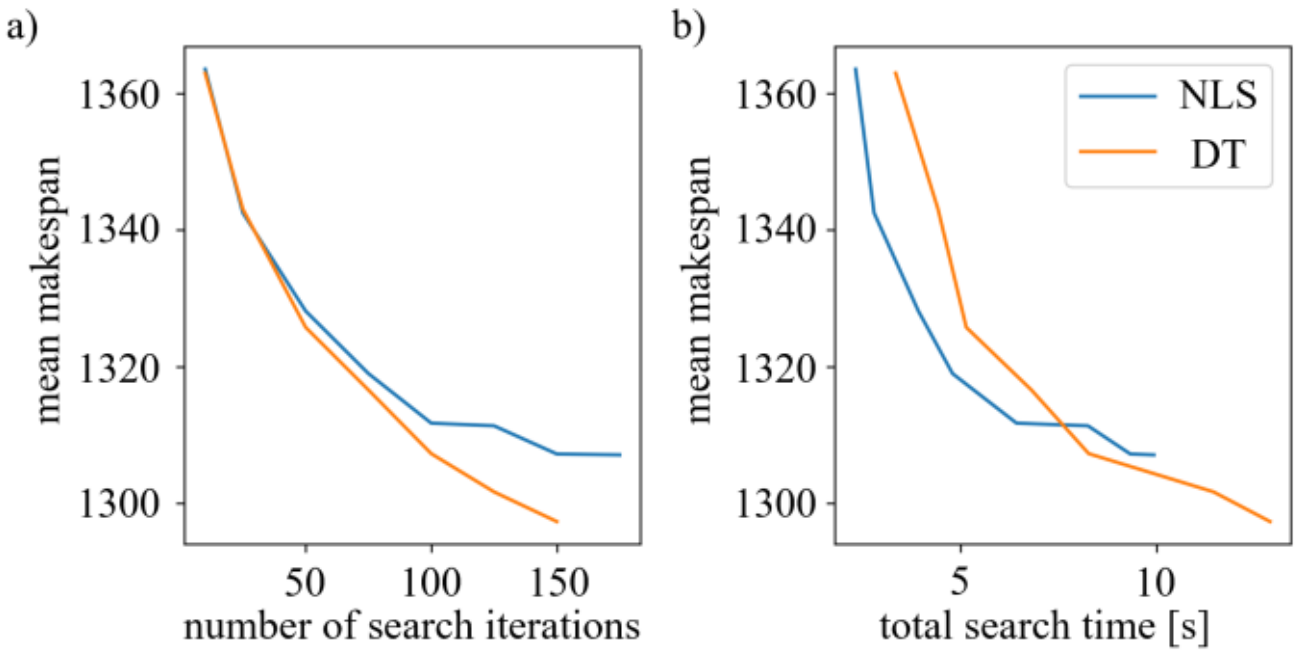

**Fig. 5**: Comparison of NLS and DT models with respect to the number of iterations (a) and the total search time in seconds (b)

The implication of longer inference times is depicted in **Fig. 5**. Plot a) on the left shows the mean makespan in relation to the number of search iterations of the teacher ($NLS_{ANP}$) and the DT model on our 15x15 JSSP instances. The gap between the models increases with a larger number of iterations, i.e., the DT model continues to improve the solution faster with more steps. Plot b) in **Fig. 5** on the right shows the same achieved mean makespans in relation to the computational time required for the complete search on our hardware (Intel Core i9-9980HK CPU @ 2.40GHz with 64 GB RAM and NVIDIA Quadro T2000 GPU). Since the teacher model performs at par with the DT for small numbers of iteration but is faster, it outperforms the DT up to search times of about 7 seconds.

However, since the teacher model converges to a worse average makespan, the DT achieves better makespans for all search times larger than seven seconds. Therefore, practically speaking, using the DT is beneficial, if search times greater than seven seconds are acceptable. It is important to note that the point at which this tradeoff occurs can vary significantly depending on the final performance of the DT and the size of the problem instances. In some cases, such as the 20x20 JSSP, the DT may not outperform the teacher model.

To compare the learned strategies between teacher and DT models, we compare the frequencies with which available actions were chosen when solving the 100 generated instances. We hypothesized that DT models achieving makespans similar to their teacher models may have learned to merely replicate the demonstrated behavior. The action frequencies of the 15x15 DT, representing the DT with the greatest relative improvement over its teacher model, is shown in **Fig. 6**a). The 20x20 and 50x20 DT models, as representatives with similar performance to their teacher models, are shown in **Fig. 6**b) and **Fig. 6**c), respectively. The available actions on the x-axis comprise the neighborhood operators CT, CET, ECET and CEI, and the perturb operator, each in combination with a reject or accept action (odd and even numbers). As one may expect based on the performance difference, the action frequencies vary between the teacher and DT model on 15x15 instances. While the frequency of taking the (CT, reject) action 0 is similar, the DT model has learned to never take the ECET actions 4 and 5, and heavily prioritizes the (CET, accept) action 3 in favor of fewer (CEI, not reject) actions 7. The action frequency distribution in 20x20 is, in comparison, much more similar between the teacher and the DT model. This, along with the similar makespans reported in TABLE 4, suggests a very similar behavior of the two models. Contrarily, 50x20 presents itself as a counterexample with similar makespans but a very different learned behavior expressed by the action frequencies in **Fig. 6**c).

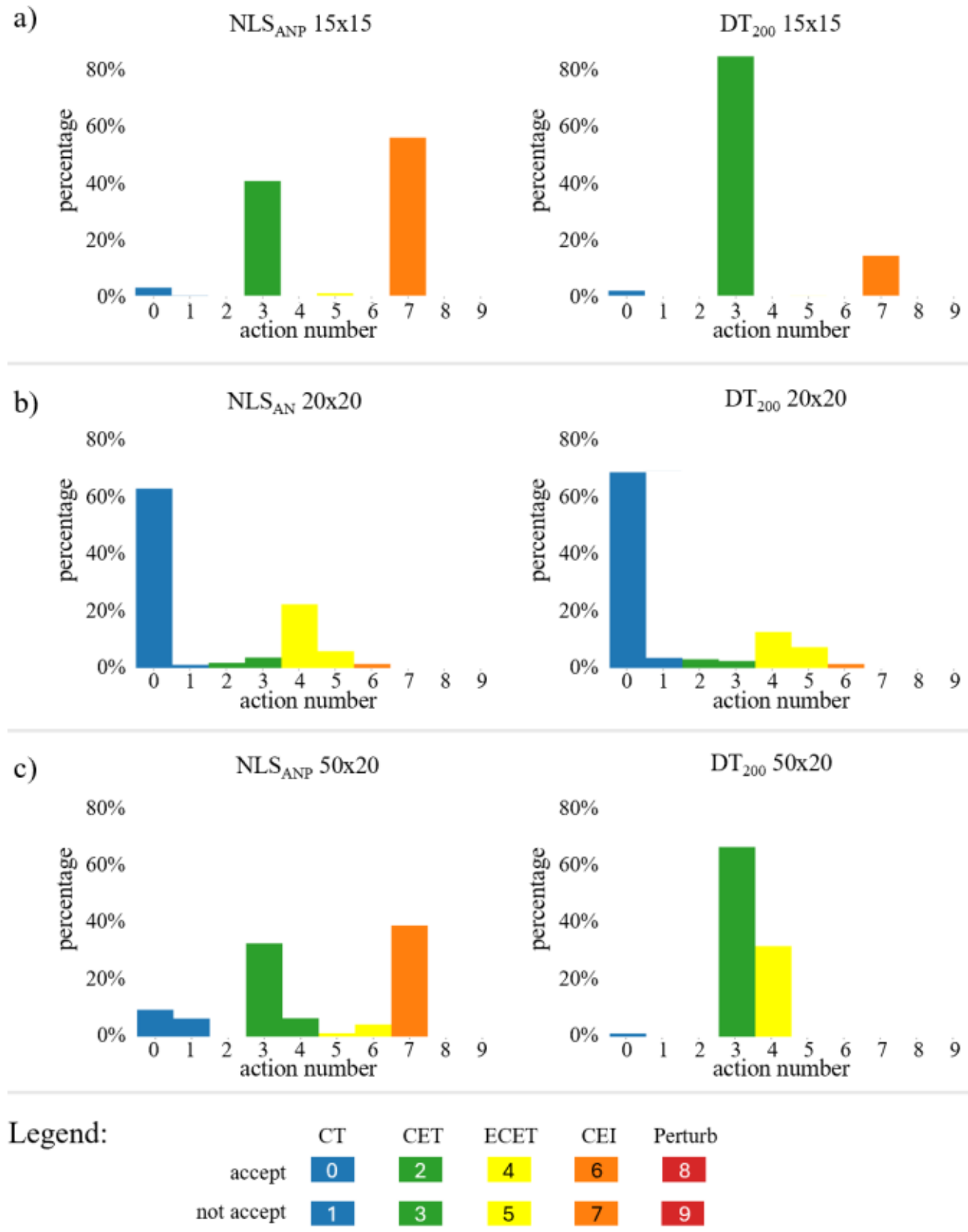

**Fig. 6**: Action frequencies of NLS and DT models

This falsifies our hypothesis that similar performance results alone indicate that the DT models are merely copying the teacher's behavior. As an interesting observation, no perturbation operation actions 8 and 9 were taken by the neither teacher nor student models that include this action ($NLS_{ANP}$ and $DT_{200}$ 15x15, $NLS_{ANP}$ and $DT_{200}$ 50x20). This is noteworthy because the ANP action sets, which include these operations, performed better than the AN action sets. This behavior is an artifact of the NLS training, not the DT training. However, it serves as an indication that the DT learns to ignore actions which have not been used by the teacher and is therefore limited in how much it can deviate from demonstrated strategies.

To analyze the influence of the return-to-go prior, we test our 15x15 $DT_{100}$ on the 100 randomly generated instances while multiplying the initial lower bound makespan $m_{lb}$ which was used as return-to-go starting point during inference with return-to-go factors in the interval [0.05, 1.75] in 0.05 increments. The interval was chosen such that the resulting return-to-go priors cover the achieved makespans of the $DT_{100}$ model. The results are shown in **Fig. 6**. We do not observe a trend in the mean makespan over different return-to-go factors. Additionally, the 95% confidence interval, that is depicted as shaded area, is much larger than the difference between the minimal and maximal values. We therefore conclude that the return-to-go has no statistically significant influence on the resulting mean makespans. Interestingly, all obtained mean makespans are smaller than those of the teacher model. This means that the DT has learned a new superior strategy independently from the return-to-go.

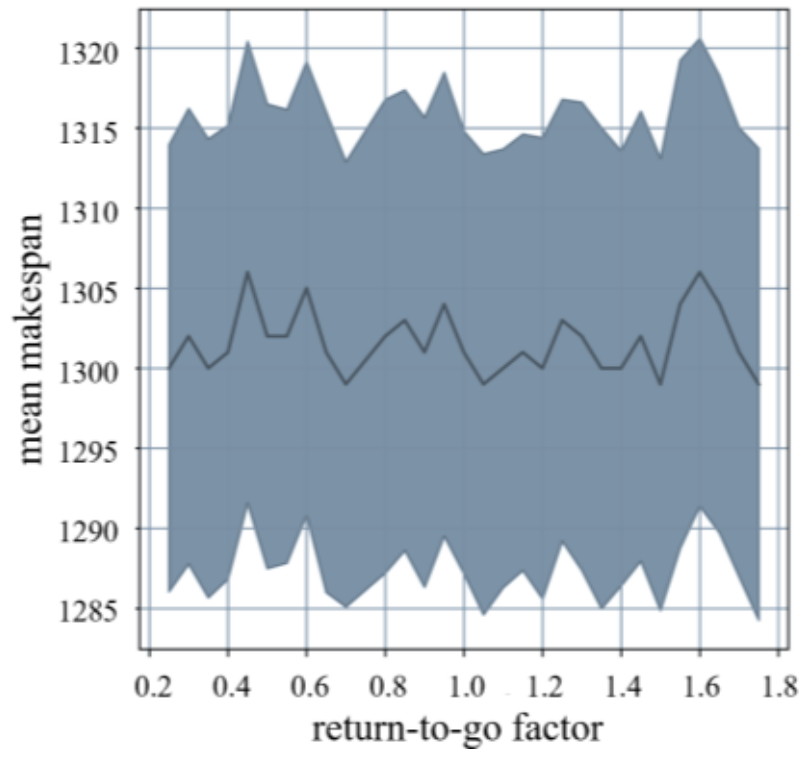

**Fig. 7**: Mean makespans of the DT on 100 15x15 instances with varying returns-to-go

## VII. Concluding Discussion and Outlook

In conclusion, we have shown that the DT can be leveraged to increase the performance of state-of-the-art NLS models on the JSSP by learning to take more effective decisions during the search. However, the benefit of using the DT varies for different problem sizes and parameterizations of the approach. We did not observe any trend that allows us to predict how much better a DT will perform a priori. From a practical perspective, this means that the benefit of using the DT may not always be given when a short search time limit applies. Upon closer examination of why the DT models outperform the NLS models, we found that their main advantage lies in the broader context of previous search steps that the DT effectively leverages through the transformer architecture. The returns-to-go on the other hand, seem to play only a minor role in reaching superior performance.

These observations motivate future work, in which we aim to push the DT towards considering the return-to-go through forced variation in the quality of solutions generated by the teacher NLS models. This could be achieved by learning from different teachers at the same time or by curriculum learning, which varies the difficulty the algorithm has with solving certain instances well, as done in [50].

## References

[1] V. Samsonov *et al.,* "Manufacturing Control in Job Shop Environments with Reinforcement Learning," in *ICAART 2021: Proceedings of the 13th International Conference on Agents and Artificial Intelligence : online streaming, February 4-6, 2021*, 2021, pp. 589–597.

[2] I. B. Park and J. Park, "Scalable Scheduling of Semiconductor Packaging Facilities Using Deep Reinforcement Learning," *IEEE Transactions on Cybernetics*, doi: 10.1109/TCYB.2021.3128075.

[3] B. Kocot, P. Czarnul, and J. Proficz, "Energy-Aware Scheduling for High-Performance Computing Systems: A Survey," *Energies*, vol. 16, no. 2, p. 890, 2023, doi: 10.3390/en16020890.

[4] M. Pinedo, *Scheduling: Theory, algorithms, and systems*: Springer International Publishing, 2016.

[5] E. Taillard, "Benchmarks for basic scheduling problems," (in en), *European Journal of Operational Research*, vol. 64, no. 2, pp. 278–285, 1993, doi: 10.1016/0377-2217(93)90182-m.

[6] J. J. van Hoorn, "The Current state of bounds on benchmark instances of the job-shop scheduling problem," *Journal of Scheduling*, vol. 21, no. 1, pp. 127–128, 2018, doi: 10.1007/s10951-017-0547-8.

[7] E. Demirkol, S. Mehta, and R. Uzsoy, "Benchmarks for shop scheduling problems," *European Journal of Operational Research*, vol. 109, no. 1, pp. 137–141, 1998, doi: 10.1016/S0377-2217(97)00019-2.

[8] T. van Ekeris, R. Meyes, and T. Meisen, "Discovering Heuristics And Metaheuristics For Job Shop Scheduling From Scratch Via Deep Reinforcement Learning," (in eng), *Proceedings of the Conference on Production Systems and Logistics : CPSL 2021*, pp. 709–718, 2021, doi: 10.15488/11231.

[9] N. Stricker, A. Kuhnle, R. Sturm, and S. Friess, "Reinforcement learning for adaptive order dispatching in the semiconductor industry," *CIRP Annals*, vol. 67, no. 1, pp. 511–514, 2018, doi: 10.1016/j.cirp.2018.04.041.

[10] P. Tassel, M. Gebser, and K. Schekotihin, "A Reinforcement Learning Environment For Job-Shop Scheduling," Apr. 2021. [Online]. Available: https://arxiv.org/pdf/2104.03760

[11] P. Tassel, B. Kovács, M. Gebser, K. Schekotihin, W. Kohlenbrein, and P. Schrott-Kostwein, "Reinforcement Learning of Dispatching Strategies for Large-Scale Industrial Scheduling," (in en), *ICAPS*, vol. 32, pp. 638–646, 2022, doi: 10.1609/icaps.v32i1.19852.

[12] C. Waubert de Puiseau, J. Peters, C. Dörpelkus, H. Tercan, and T. Meisen, "schlably: A Python framework for deep reinforcement learning based scheduling experiments," *SoftwareX*, vol. 22, p. 101383, 2023, doi: 10.1016/j.softx.2023.101383.

[13] L. Cheng, Q. Tang, L. Zhang, and Z. Zhang, "Multi-objective Q-learning-based hyper-heuristic with Bi-criteria selection for energy-aware mixed shop scheduling," *Swarm and Evolutionary Computation*, p. 100985, 2021, doi: 10.1016/j.swevo.2021.100985.

[14] R. Reijnen, Y. Zhang, Z. Bukhsh, and M. Guzek, "Learning to Adapt Genetic Algorithms for Multi-Objective Flexible Job Shop Scheduling Problems," in *Proceedings of the Companion Conference on Genetic and Evolutionary Computation*, New York, NY, USA, 2023.

[15] Z. Han, Y. Yang, and H. Ye, "A deep reinforcement learning based multiple meta-heuristic methods approach for resource constrained multi-project scheduling problem," in *2022 7th International Conference on Intelligent Computing and Signal Processing (ICSP)*, 2022, pp. 26–29.

[16] X. Chen and Y. Tian, "Learning to Perform Local Rewriting for Combinatorial Optimization," *Proceedings of the 33rd International Conference on Neural Information Processing Systems*, pp. 6281–6292, 2019. [Online]. Available: https://arxiv.org/pdf/1810.00337

[17] P. Tassel, M. Gebser, and K. Schekotihin, "An End-to-End Reinforcement Learning Approach for Job-Shop Scheduling Problems Based on Constraint Programming," *2nd Conference on Production Systems and Logistics*, pp. 614–622, 2023.

[18] A. Vaswani *et al.,* "Attention Is All You Need," *Proceedings of the 31st International Conference on Neural Information Processing Systems*, pp. 6000–6010, 2017. [Online]. Available: http://arxiv.org/pdf/1706.03762

[19] T. B. Brown *et al.,* "Language Models are Few-Shot Learners," *Advances in Neural Information Processing Systems 33*, pp. 1877–1901, 2020.

[20] L. Chen *et al.,* "Decision Transformer: Reinforcement Learning via Sequence Modeling," *Advances in neural information processing systems 34*, pp. 15084–15097, 2021.

[21] L. Perron and F. Didier, *CP-SAT*: Google, 2024. [Online]. Available: https://developers.google.com/optimization/cp/cp_solver/

[22] R. Haupt, "A survey of priority rule-based scheduling," *OR Spectrum*, vol. 11, no. 1, pp. 3–16, 1989, doi: 10.1007/BF01721162.

[23] J. Adams, E. Balas, and D. Zawack, "The Shifting Bottleneck Procedure for Job Shop Scheduling," *Management Science*, vol. 34, no. 3, pp. 391–401, 1988, doi: 10.1287/mnsc.34.3.391.

[24] N. Bhatt and N. R. Chauhan, "Genetic algorithm applications on Job Shop Scheduling Problem: A review," in *2015 International Conference on Soft Computing Techniques and Implementations (ICSCTI)*, 2015.

[25] H. R. Lourenço, O. C. Martin, and T. Stützle, "Iterated Local Search: Framework and Applications," (in en), *Handbook of Metaheuristics*, vol. 272, pp. 129–168, 2019, doi: 10.1007/978-3-319-91086-4_5.

[26] P. Hansen, N. Mladenović, J. Brimberg, and J. A. M. Pérez, "Variable Neighborhood Search," (in en), *Handbook of Metaheuristics*, vol. 272, pp. 57–97, 2019, doi: 10.1007/978-3-319-91086-4_3.

[27] R. S. Sutton and A. Barto, *Reinforcement learning: An introduction*. Cambridge, Massachusetts, London, England: The MIT Press, 2018.

[28] W. Ding *et al.,* "Learning to View: Decision Transformers for Active Object Detection," in *2023 IEEE International Conference on Robotics and Automation (ICRA)*, 2023.

[29] S. Wang, X. Chen, D. Jannach, and L. Yao, "Causal Decision Transformer for Recommender Systems via Offline Reinforcement Learning," in *Proceedings of the 46th International ACM SIGIR Conference on Research and Development in Information Retrieval*, New York, NY, USA, 2023.

[30] Yao Lai, Jinxin Liu, Zhentao Tang, Bin Wang, Jianye Hao, and Ping Luo, "ChiPFormer: Transferable Chip Placement via Offline Decision Transformer," (in en), *International Conference on Machine Learning*, pp. 18346–18364, 2023. [Online]. Available: https://proceedings.mlr.press/v202/lai23c.html

[31] J. K. Falkner, D. Thyssens, A. Bdeir, and L. Schmidt-Thieme, "Learning to Control Local Search for Combinatorial Optimization," *Falkner, Jonas K., et al. "Learning to Control Local Search for Combinatorial Optimization." Joint European Conference on Machine Learning and Knowledge Discovery in Databases*, vol. 13717, pp. 361–376, 2022, doi: 10.1007/978-3-031-26419-1_22.

[32] P. J. M. van Laarhoven, E. H. L. Aarts, and J. K. Lenstra, "Job Shop Scheduling by Simulated Annealing," *Operations Research*, vol. 40, no. 1, pp. 113–125, 1992, doi: 10.1287/opre.40.1.113.

[33] E. Nowicki and C. Smutnicki, "A Fast Taboo Search Algorithm for the Job Shop Problem," *Management Science*, vol. 42, no. 6, pp. 797–813, 1996, doi: 10.1287/mnsc.42.6.797.

[34] J. Kuhpfahl and C. Bierwirth, "A study on local search neighborhoods for the job shop scheduling problem with total weighted tardiness objective," *Computers & Operations Research*, vol. 66, pp. 44–57, 2016, doi: 10.1016/j.cor.2015.07.011.

[35] E. Balas and A. Vazacopoulos, "Guided Local Search with Shifting Bottleneck for Job Shop Scheduling," *Management Science*, vol. 44, no. 2, pp. 262–275, 1998, doi: 10.1287/mnsc.44.2.262.

[36] H. van Hasselt, A. Guez, and D. Silver, "Deep Reinforcement Learning with Double Q-Learning," *AAAI*, vol. 30, no. 1, 2016, doi: 10.1609/aaai.v30i1.10295.

[37] C. Zhang, W. Song, Z. Cao, J. Zhang, P. S. Tan, and X. Chi, "Learning to Dispatch for Job Shop Scheduling via Deep Reinforcement Learning," *Advances in Neural Information Processing Systems*, vol. 33, pp. 1621–1632, 2020.

[38] A. Corsini, A. Porrello, S. Calderara, and M. Dell'Amico, "Self-Labeling the Job Shop Scheduling Problem," Jan. 2024. [Online]. Available: http://arxiv.org/pdf/2401.11849.pdf

[39] J. Pirnay and D. G. Grimm, "Self-Improvement for Neural Combinatorial Optimization: Sample without Replacement, but Improvement," Mar. 2024. [Online]. Available: http://arxiv.org/pdf/2403.15180

[40] J. Park, J. Chun, S. H. Kim, Y. Kim, and J. Park, "Learning to schedule job-shop problems: representation and policy learning using graph neural network and reinforcement learning," *International journal of production research*, vol. 59, no. 11, pp. 3360–3377, 2021, doi: 10.1080/00207543.2020.1870013.

[41] C. Waubert de Puiseau, R. Meyes, and T. Meisen, "On reliability of reinforcement learning based production scheduling systems: a comparative survey," *Journal of Intelligent Manufacturing*, vol. 33, no. 4, pp. 911–927, 2022, doi: 10.1007/s10845-022-01915-2.

[42] C. W. d. Puiseau, C. Dörpelkus, J. Peters, H. Tercan, and T. Meisen, "Beyond Training: Optimizing Reinforcement Learning Based Job Shop Scheduling Through Adaptive Action Sampling," Jun. 2024. [Online]. Available: http://arxiv.org/pdf/2406.07325v1

[43] H. Ingimundardottir and T. P. Runarsson, "Discovering dispatching rules from data using imitation learning: A case study for the job-shop

problem," *Journal of Scheduling*, vol. 21, no. 4, pp. 413–428, 2018, doi: 10.1007/s10951-017-0534-0.

[44] A. Rinciog, C. Mieth, P. M. Scheikl, and A. Meyer, "Sheet-Metal Production Scheduling Using AlphaGo Zero," (in eng), *Proceedings of the Conference on Production Systems and Logistics : CPSL 2020*, pp. 342–352, 2020, doi: 10.15488/9676.

[45] J.-H. Lee and H.-J. Kim, "Imitation Learning for Real-Time Job Shop Scheduling Using Graph-Based Representation," in *2022 Winter Simulation Conference (WSC)*, Singapore, Dec. 2022 - Dec. 2022, pp. 3285–3296.

[46] J. K. Falkner, "NeuroLS," *GitHub*, 2024. [Online]. Available: https://github.com/jokofa/NeuroLS

[47] V. Sels, N. Gheysen, and M. Vanhoucke, "A comparison of priority rules for the job shop scheduling problem under different flow time- and tardiness-related objective functions," *International journal of production research*, vol. 50, no. 15, pp. 4255–4270, 2012, doi: 10.1080/00207543.2011.611539.

[48] D. P. Kingma and J. Ba, *Adam: A Method for Stochastic Optimization*, 2017.

[49] A. Karpathy, *minGPT*, 2020. Accessed: Jul. 25 2024.

[50] C. Waubert de Puiseau, H. Tercan, and T. Meisen, "Curriculum Learning in Job Shop Scheduling using Reinforcement Learning," *Proceedings of the Conference on Production Systems and Logistics: CPSL 2023*, pp. 34–43, 2023, doi: 10.15488/13422.